\documentclass[preprint,12pt,authoryear]{elsarticle}

\usepackage[authoryear]{natbib}
\usepackage{xcolor}
\usepackage{amssymb}
\usepackage{booktabs}
\usepackage{amsmath}
\usepackage{url}
\usepackage{float}
\begin{document}

\begin{frontmatter}



% \title{Seasonal Precipitation Forecasting in South America using Machine, Deep Learning and Explainable Artificial Intelligence} %% Article title

\title{Exploring Machine Learning, Deep Learning, and Explainable AI Methods for Seasonal Precipitation Prediction in South America}

% \affiliation[label1]{organization={},

% \affiliation[label2]{organization={},

% \author{} %% Author name

% \affiliation{organization={},%Department and Organization

\author[label1]{Matheus Corrêa Domingos\corref{cor1}}
\ead{matheus.domingos@inpe.br}

\author[label1]{Valdivino Alexandre de Santiago Júnior}
\author[label1]{Juliana Aparecida Anochi}
\author[label2]{Elcio Hideiti Shiguemori}
\author[label1]{Luísa Mirelle Costa dos Santos}
\author[label1]{Hércules Carlos dos Santos Pereira}
\author[label1]{André Estevam Costa Oliveira}

\cortext[cor1]{Corresponding author: Matheus Corrêa Domingos (matheus.domingos@inpe.br)}

\affiliation[label1]{organization={Laboratório de Inteligência ARtificial para Aplicações AeroEspaciais e Ambientais (LIAREA), Coordenação de Pesquisa Aplicada e Desenvolvimento Tecnológico (COPDT), Programa de Pós-Graduação em Computação Aplicada (PGCAP), Instituto Nacional de Pesquisas Espaciais (INPE)},
            city={São José dos Campos},
            postcode={12227-010},
            state={SP},
            country={Brasil}}

\affiliation[label2]{organization={Laboratório de Inteligência ARtificial para Aplicações AeroEspaciais e Ambientais (LIAREA), Instituto de Estudos Avançados (IEAv), Programa de Pós-Graduação em Computação Aplicada (PGCAP)},
            city={São José dos Campos},
            postcode={12228-001},
            state={SP},
            country={Brasil}}

%% Abstract
\begin{abstract}
Forecasting meteorological variables is challenging due to the complexity of their processes, requiring advanced models for accuracy. Accurate precipitation forecasts are vital for society. Reliable predictions help communities mitigate climatic impacts. Based on the current relevance of artificial intelligence (AI), classical machine learning (ML) and deep learning (DL) techniques have been used as an alternative or complement to dynamic modeling. However, there is still a lack of broad investigations into the feasibility of purely data-driven approaches for precipitation forecasting. This study aims at addressing this issue where different classical ML and DL approaches for forecasting precipitation in South America, taking into account all 2019 seasons, are considered in a detailed investigation. The selected classical ML techniques were Random Forests and extreme gradient boosting (XGBoost), while the DL counterparts were a 1D convolutional neural network (CNN 1D), a long short-term memory (LSTM) model, and a gated recurrent unit (GRU) model. Additionally, the Brazilian Global Atmospheric Model (BAM) was used as a representative of the traditional dynamic modeling approach. We also relied on explainable artificial intelligence (XAI) to provide some explanations for the models behaviors. LSTM showed strong predictive performance while BAM, the traditional dynamic model representative, had the worst results. Despite presented the higher latency, LSTM was most accurate for heavy precipitation. If cost is a concern, XGBoost offers lower latency with slightly accuracy loss. The results of this research confirm the viability of DL models for climate forecasting, solidifying a global trend in major meteorological and climate forecasting centers.

%LSTM showed strong predictive performance, with MSE of 1.28 (summer), 0.91 (autumn), 0.83 (winter), 0.43 (spring), and latency of 2,736.8 ms, 2,843.1 ms, 28,266.0 ms, and 2,765.9 ms, respectively. BAM had the worst results. LSTM, despite higher latency, was most accurate for heavy precipitation. If cost is a concern, XGBoost offers lower latency with slight accuracy loss.
\end{abstract}


% \begin{highlights}
% \end{highlights}

%% Keywords

\end{frontmatter}


%% Use \subsubsection, \paragraph, \subparagraph commands to 
%% start 3rd, 4th and 5th level sections.
%% Refer following link for more details.
%% https://en.wikibooks.org/wiki/LaTeX/Document_Structure#Sectioning_commands

%%       \citet{<label>} ==> Lamport (1994)
%%       \citep{<label>} ==> (Lamport, 1994)
%%




\section{INTRODUCTION}

With climate change intensifying, it becomes increasingly essential to obtain consistent and reliable forecasts to reduce the impacts of extreme weather events. These phenomena are transforming ecosystems, altering the availability of water resources and affecting strategic sectors such as agriculture and urban infrastructure \citep{angelotti2019acoes, marengo2024maior, paiva2024cooperacao}.

Precipitation forecasting, in particular, represents a complex challenge, especially in accurately estimating the magnitude of this meteorological phenomenon. Atmospheric forecasting can be categorized into weather forecasting and climate forecasting. While weather forecasting is based on the initial conditions of the atmosphere to anticipate short-term events, climate forecasting evaluates long-term atmospheric patterns, considering aggregated statistical variables \citep{reboita2010regimes}.

Previous studies investigated the seasonal climate prediction of precipitation and air temperature at 2 meters in Brazil, using global models such as the Climate Forecast System Version 2 (CFSv2), operated by the Center for Weather Forecasting and Climate Studies (CPTEC), as well as the regional model RegCM4. The performance of these models was analyzed using metrics such as bias and Willmott's concordance index, highlighting significant seasonal discrepancies between predicted and observed values, with a tendency to underestimate or overestimate data \citep{reboita2018previsao}.

Traditional climate prediction models, based on solving partial differential equations, face challenges in capturing the complex interactions between the atmosphere and the Earth's surface. Although they are widely used, their ability to represent localized events and seasonal variations still has limitations \cite{holton2013introduction}.

In this context, technological advances and artificial intelligence (AI) have revolutionized the analysis of natural phenomena, providing more efficient methods for identifying patterns in large volumes of data. Machine learning (ML) and deep learning (DL) techniques have demonstrated great potential for improving the accuracy of weather forecasts \citep{oliveira2023precipitation}.
Recent studies have explored different approaches based on neural networks and ML for precipitation prediction. Ramirez, De Campos Velho and Ferreira (2005) used artificial neural networks (ANNs) to establish a non-linear mapping between the output of the regional Eta model and surface precipitation data, aiming at quantitative predictions for the state of São Paulo. \cite{freitas2019deep} highlighted the use of decision trees in predicting short-scale deep convection in the Metropolitan Region of Rio de Janeiro. \cite{mahajan2022prediction} applied Random Forests to predict rainfall occurrence in Australia, comparing the results with other models such as Support Vector Machine (SVM), Naïve Bayes and K-nearest neighbors (KNN).

In \cite{anochi2020neural}, a DL model for seasonal climate precipitation prediction was proposed, using data from the Global Precipitation Climatology Project (GPCP). The study employed a multilayer perceptron neural network (MLP) optimized by the Multiparticle Collision Algorithm (MPCA), demonstrating the effectiveness of this approach in seasonal forecasting. \cite{gibson2021training} used Random Forests to predict seasonal precipitation in the western United States, exploring interpretability methods to analyze the underlying physical processes in the data.

\cite{rossatto2023recurrent} uses Recurrent Convolutional Neural Networks (RCRNNs) for short-term weather forecasting (nowcasting) with weather radar images in Brazil. The study demonstrates that these networks have great potential in improving predictions of extreme events, such as storms, achieving up to 90\% accuracy for 60-minute forecasts.

Despite these previous studies, broad investigations regarding the feasibility of purely data-driven approaches, such as ML and DL algorithms/models, to precipitation forecasting are still lacking. In other words, it is necessary to explore different ML/DL techniques, see what are their advantages and limitations and thus provide evidence that certain approaches are more suitable than others for seasonal precipitation forecasting. This study then aims at addressing this issue where different classical ML and DL techniques for predicting precipitation in South American, considering all 2019 seasons, are taking into account. The selected classical ML techniques were Random Forests \cite{breiman2001random} and extreme gradient boosting (XGBoost) \citep{niazkar2024applications} while DL counterparts were a convolutional neural network 1D (CNN 1D) \citep{cacciari2024hands}, a long short-term memory (LSTM) model (Hochreiter $\&$ Schmidhuber, 1997), and a gated recurrent unit (GRU) model \citep{cho2014learning}. The research employs monthly GPCP observations to generate predictions, and evaluated the AI approaches considering metrics such as mean squared error (MSE), coefficient of determination (R²), probability of detection (POD), false alarm rate (FAR), and latency in the test set. Thus, we evaluated not only the general benefit of each model but also its cost. Moreover, the Brazilian Global Atmospheric Model (BAM) \citep{moura2021evaluation} was used as a representative of the traditional dynamic modeling approach.

The main contribution of this study is precisely such a broad investigation of different classical ML and DL approaches to a challenging problem, climate precipitation prediction, considering all seasons of one year and also a traditional dynamic model within such a comparison. Furthermore, SHapley Additive exPlanations (SHAP) analyses were carried out for the autumn and spring seasons to provide insights into model explainability and feature importance \citep{lundberg2017unified}.

\section{RELATED WORK}
In this section, we present some relevant related studies highlighting their contributions and limitations.

\cite{anochi2021machine} investigated the seasonal climate prediction of precipitation in South America for the years 2018 and 2019, using data from the GPCP with a spatial resolution of $2.5^\circ \times 2.5^\circ$. The authors compared the performance of an MLP optimized by the MPCA with the model without MPCA and with the BAM model (widely used in operational climate forecasts). The results indicated that the MPCA improved the accuracy of the MLP, but still had limitations in some regions, as indicated by the MSE and other evaluation metrics.

Seeking to improve these results, \cite{monego2022south} explored the use of XGBoost, a decision tree-based machine learning model, applying Optuna for hyperparameter optimization. The study focused on improving the accuracy of seasonal forecasts for 2018 and 2019. The results were compared with those of MLP and BAM, demonstrating significant gains in forecast accuracy, especially in reducing MSE.

More recently, \cite{anochi2024precipitation} extended the approach to monthly precipitation forecasting in 2023 and incorporated a drought analysis. The research used data from the GPCP (1983-2023) and employed an MLP model to generate predictions, comparing the results with monthly data provided by the North American Multi-Model Ensemble (NMME). Although the metrics indicated satisfactory performance, the MLP had difficulties in capturing extreme variations and more complex regional patterns.

In the study by \cite{oliveira2023precipitation}, several works on precipitation forecasting were analyzed, evaluating model performance based on root mean squared error and mean squared error metrics. The results indicated that the XGBoost and BLSTM-GRU models (a combination of Bidirectional Long Short-Term Memory Recurrent Neural Network and Gated Recurrent Unit) showed outstanding performance, with root mean squared error ranging from 0.85 to 1.70 for XGBoost and mean squared error of 0.0075 for BLSTM-GRU. The presence of extreme events in the data proved useful for capturing complex patterns with higher accuracy. Additionally, it was observed that, depending on the temporal scale and model performance, the complexity of the evaluation can be reduced, facilitating the analysis of results. In this context, both ML and DL models proved to be effective for precipitation forecasting.

\cite{slater2023hybrid} propose a hybrid approach to climate forecasting, combining physics-based models rooted in climate dynamics with ML and DL to enhance precipitation and other hydro-meteorological variable predictions. A notable application of this approach is in extreme precipitation forecasting, where hybrid models implemented with UKMO GloSea5 and ECMWF, using a serial structure, demonstrated high performance. However, some limitations were identified, including the need to align ML and DL models with physical processes to improve result interpretability, the difficulty of these approaches in recognizing novel patterns not present in the training data, the dependence on large datasets for effective performance, and the need for continuous optimization and calibration to ensure reliable long-term forecasts.

Although these studies have promoted advances in precipitation prediction, the models used still face challenges in capturing seasonal patterns and predicting extreme events. Therefore, this research proposes exploring a wider range of deep neural networks and classical ML techniques, aiming to improve the representation of climate patterns and provide more accurate and robust forecasts for South America.

\section{MATERIALS AND METHODS}
% Citations to standard references in text should consist of the name of the
% author and the year of publication, for example, \citet{Becker+Schmitz2003} or
% \citep{Becker+Schmitz2003} using the appropriate $\backslash$citet\ or
% $\backslash$citep commands, respectively. A variety of citation formats can
% be used with the natbib package; however, the AMS prefers that authors use only the $\backslash$citet\ and
% $\backslash$citep commands. References should be entered in the references.bib file. For a thorough
% discussion of how to enter references into the references.bib database file
% following AMS style, please refer to the \textbf{AMS\_RefsV5.pdf} document
% included in this package.

% \textcolor{red}{Matheus, aqui é necessário indicar quais variáveis pertencem ao NCEP e quais pertencem ao GPCP.} - Feito

Table 1 presents the climatic variables used in this study along with their respective units. The historical data were provided by GPCP, covering the period from January 1980 to March 2020, with a spatial resolution of $2.5^\circ \times 2.5^\circ$ \citep{inmetinpe2024prognostico}. The GPCP dataset provides observational precipitation data \citep{adler2018global}, whereas the other climate variables were obtained from the NCEP-NCAR Reanalysis 1 dataset \citep{kalnay1996ncep}.

\begin{table}[h!]
\centering
\begin{tabular}{|l|l|l|}
\hline
\textbf{Variables} & \textbf{Units} & \textbf{Font}\\
\hline
Sea level pressure & milibares (mb) & NCEP-NCAR\\ \hline
Air temperature (surface) & °C & NCEP-NCAR \\ \hline
Temperature at 850 hPa & °C & NCEP-NCAR \\ \hline
Specific humidity at 850 hPa & grams/kg & NCEP-NCAR \\ \hline
Southern wind component at 850 hPa & m/s & NCEP-NCAR\\ \hline
Zonal wind component at 500 hPa & m/s & NCEP-NCAR\\ \hline
Zonal wind component at 850 hPa & m/s & NCEP-NCAR\\ \hline
Precipitation & mm/day & GPCP\\
\hline
\end{tabular}
\caption{List of variables and their respective units.}
\label{tab:variables_units}
\end{table}

%The entire datatet ... 

The entire dataset contains 24 longitude points and 28 latitude points, resulting in a total of 672 points per year, and occupies 13,963,674 bytes.

\begin{figure}[H]
    \centering
    \includegraphics[width=25pc,angle=0]{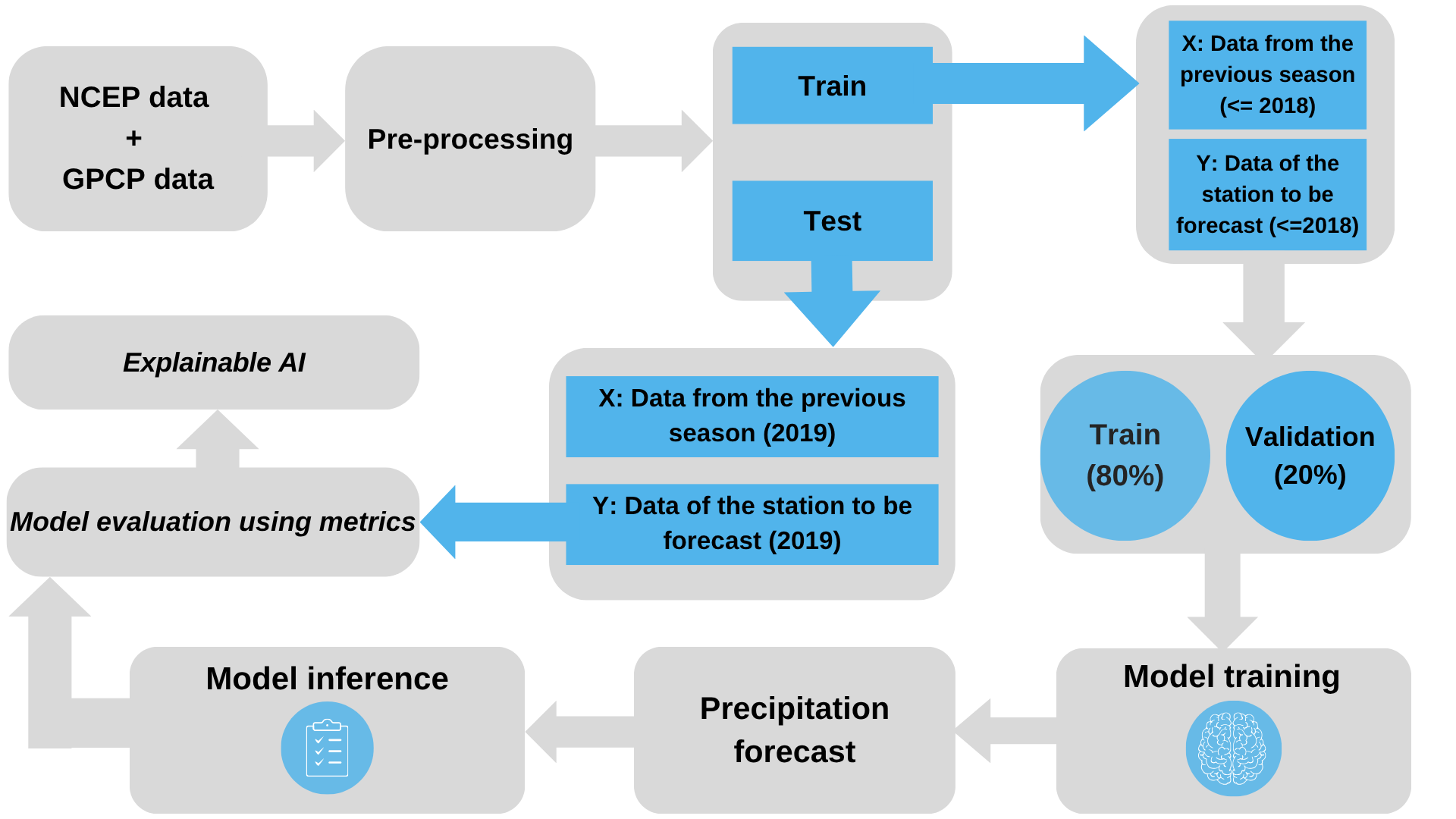}
    \caption{The methodology created for climate precipitation forecasting.}
    \label{fig:methodology}
\end{figure}

Figure \ref{fig:methodology} illustrates the flowchart of the methodology developed for this research. This diagram represents the steps taken in data processing and model development, highlighting the integration of observational and reanalysis datasets.

The data classification follows the conventional climatological definition of seasons, where Summer (DJF) encompassing December, January and February; Autumn (MAM) with March, April and May; Winter (JJA) including June, July and August; and Spring (SON) covering September, October and November. For each season, we used data from the previous season to predict conditions for the next season, applying a holdout split that divided the dataset into 80\% for training, 20\% for validation, and reserved the 2019 data for testing.

%For each season, we used data from the previous season to predict conditions for the next season, dividing the dataset into 80\% for training, 20\% for validation, and reserving the 2019 data for testing \textcolor{red}{citar a tecnica holdout}. -Feito

As for training, data were normalized using the Z-score technique, which adjusts values to a mean of 0 and standard deviation of 1, eliminating scale discrepancies between variables. This standardization not only makes data comparable, but also accelerates model convergence by improving training stability and efficiency. Equation 1 presents the formula used for this normalization.

\begin{equation}
z = \frac{X - \mu}{\sigma}
\end{equation}

where $z$ is the normalized value, X is the original value, $\mu$ is the mean of the variable and $\sigma$ is the standard deviation of the variable.

\begin{figure}[H]
    \centering
    \includegraphics[width=0.7\linewidth]{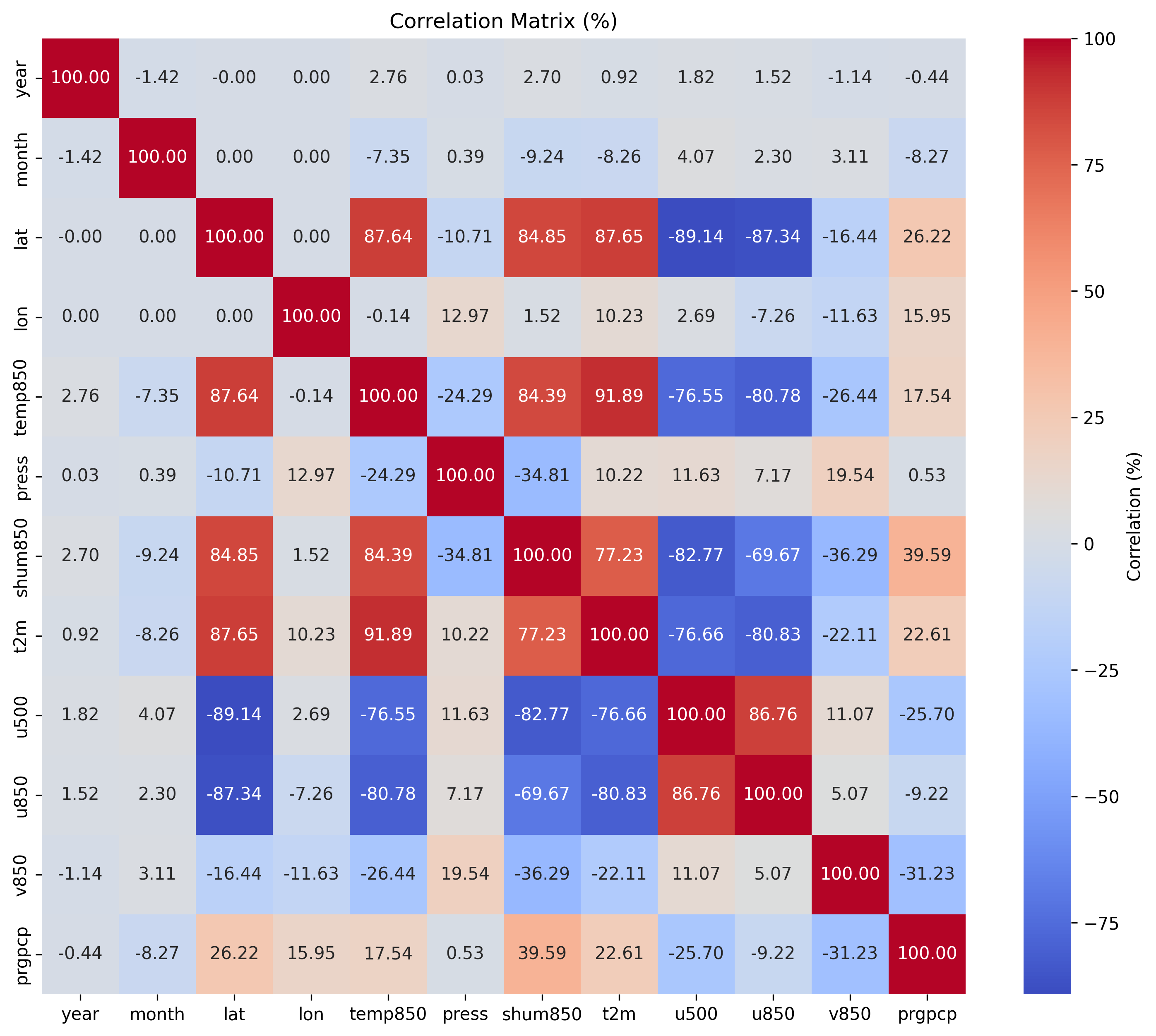}
    \caption{Correlation Matrix}
    \label{fig:correlation_matrix}
\end{figure}

A sensitivity analysis was performed to see if some variables could be discarded as they could potentially harm rather than improve the learning of the ML and DL models. Thus, variables such as pressure at surface level and air temperature at 850 hPa were removed, as they showed a high correlation (as shown in Figure \ref{fig:correlation_matrix}) with other more relevant variables, helping to eliminate redundancies and reduce multicollinearity. This step was essential to improve the efficiency of predictive models. In Figure \ref{fig:correlation_matrix}, the data correlation matrix is shown. This analysis also served as a basis for a subsequent explainability analysis of the models, evaluating which features contributed most to the predictions \citep{malakouti2025leveraging, ibebuchi2025uncertainty}.

% \textcolor{red}{Pelo nome da variável de precipitação, acredito que você tenha utilizado a versão 3.2, e não a versão 2.3 - Verificar o dataset.} - É a 2.3

Tables \ref{tab:model_params}, which presents the fixed hyperparameters of the deep learning models, \ref{tab:xgb_params}, which lists the hyperparameters for the XGBoost model, and \ref{tab:rf_params}, which contains the hyperparameters used in the Random Forest model, present the adjusted hyperparameters, detailing the configurations adopted to optimize the performance of each evaluated model.

% Based on this initial analysis, the hyperparameters of the ML and DL models were empirically tuned through experimental optimization to maximize performance. Tables \ref{tab:model_params} with hiperparametros fixos dos modelos de deep learning, \ref{tab:xgb_params} com hiperparametros para o modelo xgboost and \ref{tab:rf_params} com os hiperparametros usandos no modelo random forest below present the adjusted hyperparameters, detailing the settings used to optimize the performance of each model evaluated. 

\begin{table}[H]
\centering
\begin{tabular}{|c|c|}
\hline
\textbf{Parameters} & \textbf{Value} \\ \hline
Number of hidden layers & 5 \\ \hline
Epochs & 1000 \\ \hline
Activation function & ReLU \\ \hline
Batch size & 96 \\ \hline
Patience & 100 \\ \hline
Neurons & [256, 128, 64, 32, 64, 128, 256] \\ \hline
Optimizer & AdamW \\ \hline
Learning rate & 0.001 \\ \hline
\end{tabular}
\caption{Configuration and hyperparameters of the deep learning models.}
\label{tab:model_params}
\end{table}

\begin{table}[H]
\centering
\begin{tabular}{|l|l|}\hline
\textbf{Parameter} & \textbf{Value} \\ \hline
Learning rate & 0.1 \\ \hline
Maximum depth & 6 \\ \hline
Minimum sample weight & 1 \\ \hline
Subsample & 0.8 \\ \hline
Colsample by tree & 0.8 \\ \hline
Gamma & 0 \\ \hline
Number of processes (\texttt{n\_jobs}) & -1 \\\hline
\end{tabular}
\caption{Hyperparameters used for the XGBoost model.}
\label{tab:xgb_params}
\end{table}

\begin{table}[H]
\centering
\begin{tabular}{|l|l|}\hline
\textbf{Parameter} & \textbf{Value} \\ \hline
Number of trees & 1000 \\ \hline
Maximum depth & None \\\hline
\end{tabular}
\caption{Parameters used in the Random Forest model.}
\label{tab:rf_params}
\end{table}

\begin{figure}[H]
    \centering
    \includegraphics[width=1.0\linewidth]{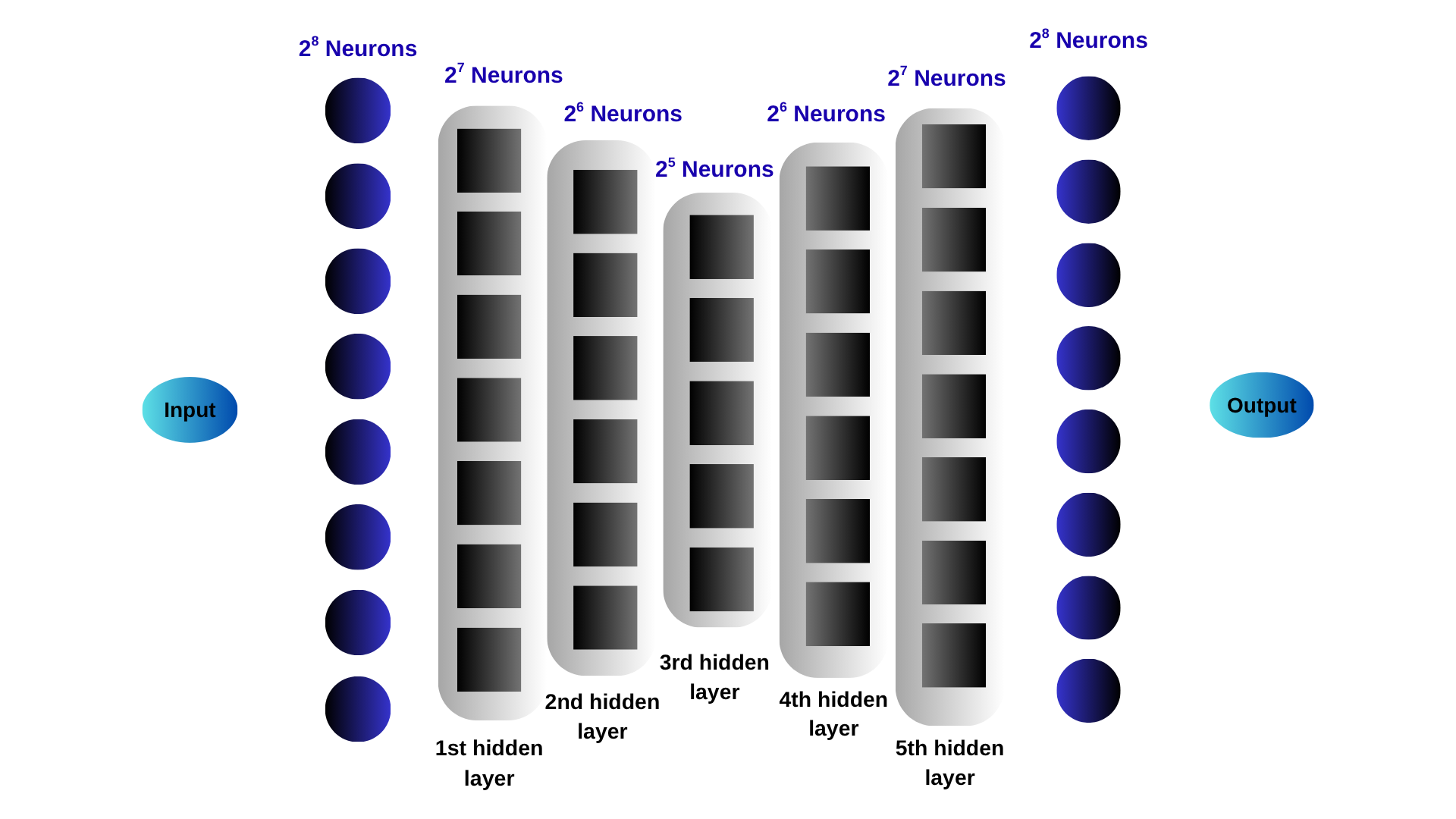}
    \caption{Neuron configuration by layers, where the number of neurons follows powers of 2.}
    \label{fig:layers}
\end{figure}

Figure~\ref{fig:layers} presents a layer configuration following powers of two: the input layer contains $2^8$ neurons; the first hidden layer has $2^7$; the second, $2^6$; the third, $2^5$; the fourth returns to $2^6$; and the fifth hidden layer contains $2^7$, totaling five hidden layers. The output layer concludes with $2^8$ neurons.

% This entire process also served as a basis for a subsequent explainability analysis, which evaluated the contribution of each feature to the model predictions, helping to identify the most relevant variables and further enhance model interpretability.

\subsection{Precipitation Prediction Approaches}

This section briefly summarizes the selected ML and DL models as well as the BAM model for precipitation forecasting.

\subsubsection{Random Forests}

Random Forests is a model composed of multiple decision trees, where each tree is trained independently with random subsets of the data and variables. This approach allows you to capture different patterns in the dataset, making the model more robust and reducing the risk of overfitting. At the end of the process, the individual outputs of the trees are combined by majority voting for classification and by averaging the predictions for regression, resulting in an efficient model for regression, classification and prediction tasks \citep{biau2012analysis}. 

\subsubsection{XGBoost}

XGBoost is a model based on multiple decision trees combined with gradient descent for optimization. Trees are generated from random subsets, where the first tree serves as the basis for the following ones. Each new tree adjusts the residuals from the previous prediction, gradually refining the model by minimizing the difference between predicted and actual values. Gradient descent is used to minimize the error function, making the process more efficient and improving the predictive capacity of the model \citep{chen2016xgboost}.

\subsubsection{LSTM}

LSTM is a recurrent neural network model that was developed to capture long-term dependencies in sequential data through a memory cell, managed by three gates. The forget gate determines what information should be discarded or kept based on the state of the memory, while the input gate updates the memory with new relevant information. Finally, the output gate selects which data will be propagated to the next stage, ensuring efficient information processing \citep{staudemeyer2019understandinglstmtutorial}.

\subsubsection{GRU}

GRU is a simplified version of LSTM that maintains high efficiency in results. This neural network has two main gates: the reset gate, which controls the amount of information to be preserved in memory, and the update gate, responsible for balancing the influence of the previous state on the new state, ensuring a smooth transition between information \citep{chung2014empirical}.

\subsubsection{CNN 1D}

1D CNN extracts features along a single dimension, capturing relevant patterns in the data. These features are then used to train an MLP model, which can be applied to regression, classification, or prediction tasks. This feature extraction process, combined with subsequent modeling, optimizes network performance, making the solution more efficient and accurate \citep{kiranyaz2021survey}.

\subsubsection{BAM}

BAM is a model used for global weather and climate prediction, characterized as a 3D spectral model with a resolution of 20 km and 96 vertical layers. It employs a semi-implicit semi-Lagrangian scheme for temporal integration. BAM generates initial conditions for weather forecasting, boundary conditions for regional models, as well as seasonal and long-term climate forecasts.

BAM incorporates different schemes to represent essential atmospheric processes. For the surface layer, the IBIS v.2.6 model, adapted and implemented by CPTEC, is used. Cloud microphysics is addressed using the Morrison scheme, which considers double momentum volume. Radiation and cloud properties are simulated by The Rapid Radiative Transfer Model (RRTMG), responsible for short and long wave radiation. Finally, the shallow convection scheme in BAM follows the methodology proposed by Park and Bretherton \citep{figueroa2016brazilian}.

\subsection{Metrics}

To evaluate the performance of the models, MSE, R², POD, FAR, and latency were calculated based on the test set. For the BAM model, latency was not considered.

The formulas for the mentioned metrics are presented below:

\begin{equation}
\text{MSE} = \frac{1}{n} \sum_{i=1}^{n} (y_i - \hat{y}_i)^2
\end{equation}

\begin{equation}
R^2 = 1 - \frac{\sum_{i=1}^{n} (y_i - \hat{y}_i)^2}{\sum_{i=1}^{n} (y_i - \bar{y})^2}
\end{equation}

\begin{equation}
\text{POD} = \frac{TP}{TP + FN}
\end{equation}

\begin{equation}
\text{FAR} = \frac{FP}{TP + FP}
\end{equation}

where \( \hat{y} \) is the prediction made by the model, \( y \) is the observed value provided by GPCP, and \( \bar{y} \) is the mean of the observed values. \( \text{TP} \) denotes the true positives, \( \text{FN} \) the false negatives, and \( \text{FP} \) the false positives. POD and FAR are used to evaluate the models' ability to identify intense precipitation events, defined as those above the 95\textsuperscript{th} percentile of data observed by GPCP, analyzed separately for each season of the year in 2019. Latency represents the model's inference time on the test set.

In general, the MSE is the most interesting metric, as it is sensitive to extreme values which makes it fundamental for rainfall forecasting, especially in heavy rain events. While a higher \( R^2 \) indicates a better explanatory capacity of the model, the MSE directly measures the accuracy of the forecasts and penalizes larger errors more significantly. In the case of POD, high values are desirable as they indicate better detection of heavy rainfall, while a lower FAR is preferable as it represents fewer false alarms. However, if POD and FAR are very close, the result is undesirable, as it suggests a high number of erroneous forecasts.

For a visual assessment of the forecasting errors, where the locations at which the models underestimated or overestimated the seasonal mean rainfall are examined, error maps will be generated using the following formula:

% For a visual analysis of the forecasting errors, in which the points were the models underestimated or overestimated the average rainfall for each season are analyzed, error maps will be generated based on the formula:

\begin{equation}
    Error_{maps}(lat,lon) = F_{prev}(lat, lon) - F_{obs}(lat,lon)
\end{equation}

where \( Error_{maps}(lat,lon) \) represents the error at each grid point, \( F_{prev}(lat, lon) \) corresponds to the predicted value at each grid point, and \( F_{obs}(lat,lon) \) corresponds to the observed value, obtained by combining the GPCP and NCEP data.

The experiments were performed partly on the SDumont computer \cite{IDeepS} and partly in the BDC-Lab environment \cite{Queiroz2024BDCLab}.

\section{Results}

In this section, we present and analyze the results of precipitation predictions for each season of the year (2019), comparing the observed values with those generated by the models, as well as the corresponding error maps. Additionally, a dedicated subsection presents an explainability analysis of the AI models for the autumn and spring seasons, using SHAP values to interpret the relative importance of climatic variables and to better understand the models’ predictive behavior during these transition periods.

\subsection{Summer}

Summer in the Southern Hemisphere is characterized by high temperatures due to the Earth's tilt in relation to the Sun, which results in more intense exposure to solar radiation. This period favors an increase in precipitation, wind intensity and electrical discharges. The South Atlantic Convergence Zone (ZCAS) exerts influence on the Southeast and Central-West regions, while the Northeast and North are impacted by the Intertropical Convergence Zone (ZCIT), which contributes to an increase in precipitation rates in these areas. This is pointed out by the Summer Climate Prognosis \citep{inmetinpe2024prognostico} and is corroborated by \cite{schneider2014migrations}.

\begin{figure}[H]
    \centering
    \includegraphics[width=1.0\linewidth]{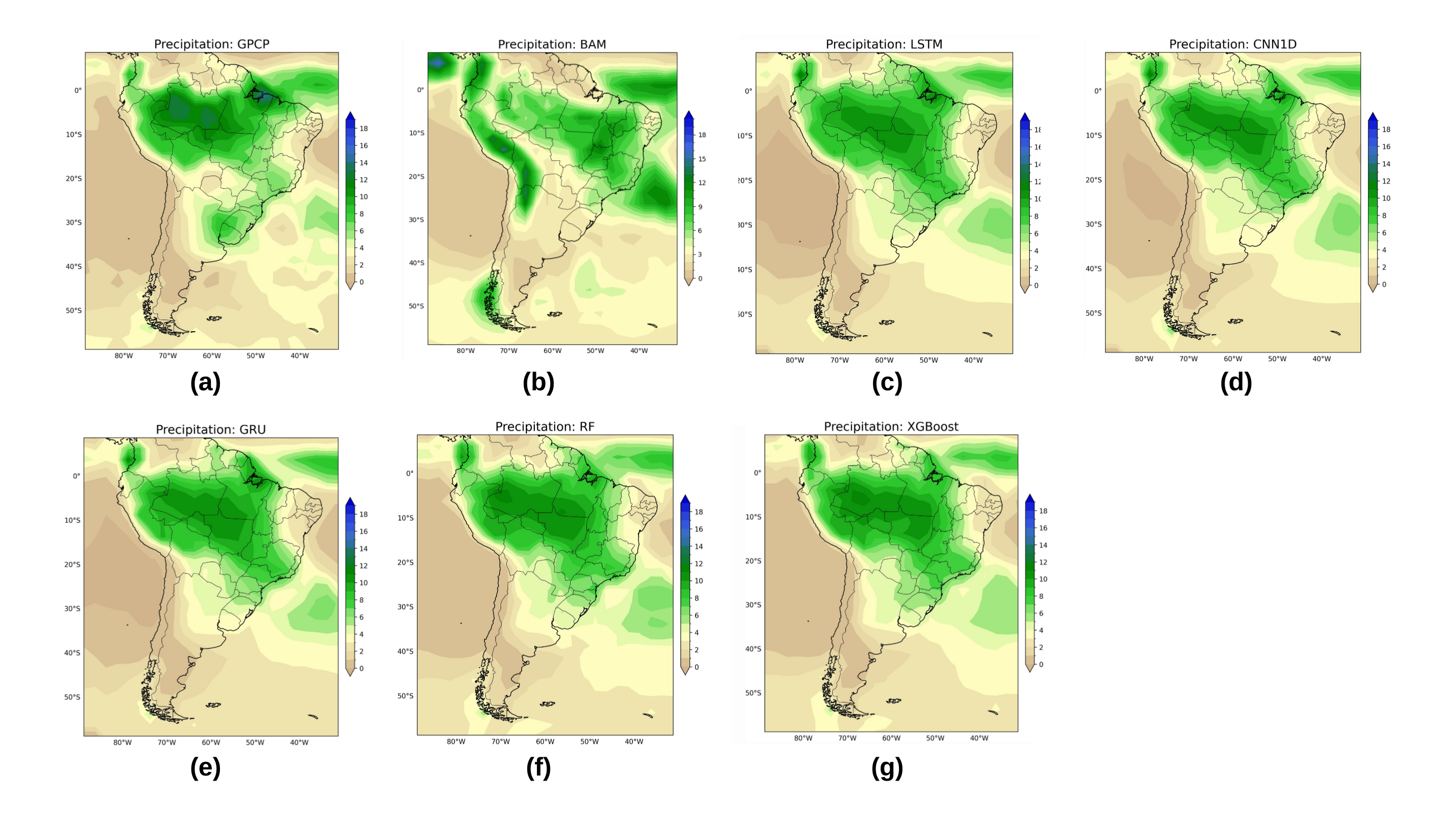}
    \caption{Maps of observed and predicted precipitation for the summer of 2019. (a) precipitation observed by GPCP, (b) precipitation by the BAM model, (c) precipitation estimated by the LSTM model, (d) precipitation estimated by the CNN1D model, (e) precipitation estimated by the GRU model, (f) precipitation estimated by the RF model and (g) precipitation estimated by the XGBoost model.}
    \label{fig:summer-mapsprec}
\end{figure}

\begin{figure}[H]
    \centering
    \includegraphics[width=0.4\linewidth]{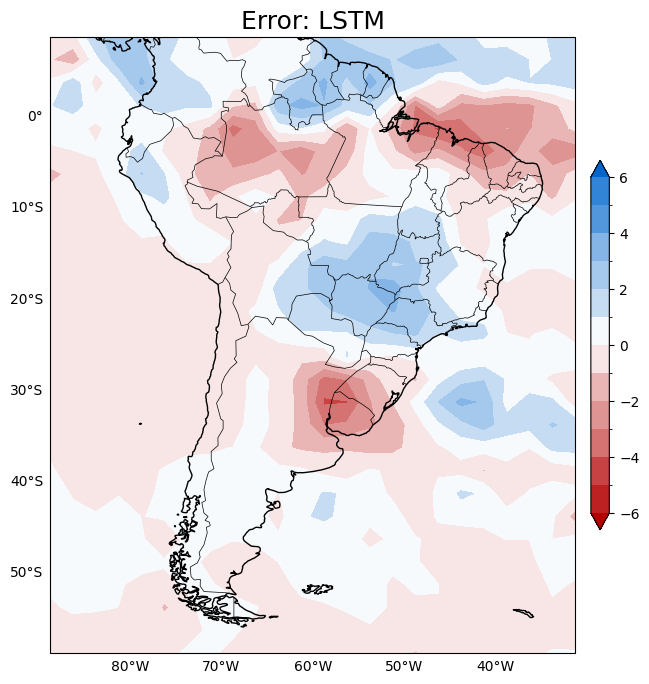}
    \caption{Error map of the best model for summer season}
    \label{fig:summer-errorprec}
\end{figure}

Figure \ref{fig:summer-mapsprec} presents the maps of precipitation observed by the GPCP and the precipitation predicted by each model during the summer of 2019. Table \ref{tab:summer} summarizes the results of the evaluation metrics for each model, while Figure \ref{fig:summer-errorprec} highlights the error associated with the best performing model.

In comparison with the GPCP observation shown in Figure \ref{fig:summer-mapsprec}, the BAM model (b) maintains a similar pattern, but overestimates precipitation in the Amazon and southern Brazil. Neural network models, such as LSTM (c), CNN1D (d) and GRU (e), capture the rainfall distribution well, but they smooth out some transitions and, in the case of GRU, underestimate areas of higher precipitation. The models based on decision trees, RF (f) and XGBoost (g), present a smoother pattern, reducing extremes and being able to overestimate semi-arid areas. An important point is that DL models tend to better represent the spatial variations of precipitation, while RF and XGBoost may have lower accuracy in areas of heavy rainfall. Furthermore, BAM, as a physical climate model, exhibits differences in the location and intensity of precipitation.

% Compared to the GPCP in figure \ref{{fig:summer-mapsprec}}, the BAM model (b) maintains a similar pattern, but overestimates precipitation in the Amazon and southern Brazil. Neural network models, such as LSTM (c), CNN1D (d) and GRU (e), capture the rainfall distribution well, but they smooth out some transitions and, in the case of GRU, underestimate areas of higher precipitation. The models based on decision trees, RF (f) and XGBoost (g), present a smoother pattern, reducing extremes and being able to overestimate semi-arid areas. An important point is that DL models tend to better represent the spatial variations of precipitation, while RF and XGBoost may have lower accuracy in areas of heavy rainfall. Furthermore, BAM, as a physical climate model, exhibits differences in the location and intensity of precipitation.

\begin{table}[H]
\centering
\caption{Results of the models with what was observed in the GPCP. Best values for each metric are in bold.}
\vspace{0.2cm}
\begin{tabular}{|l| l| l| l| l| l|}\hline
Model & Latency (ms) & MSE & R$^2$ & POD & FAR \\\hline
CNN 1D        & 7,146.194   & 1.39 & 0.84 & 0.50 & \textbf{0.23} \\\hline
LSTM          & 2,736.7680  & \textbf{1.28} & \textbf{0.85} & 0.59 & 0.35 \\\hline
GRU           & 2,960.3510  & 1.41 & 0.84 & 0.59 & 0.39 \\\hline
Random Forest & 2,563.9410  & 1.45 & 0.83 & 0.53 & 0.28 \\\hline
XGBoost       & \textbf{3.6967}      & 1.53 & 0.83 & \textbf{0.62} & 0.30 \\\hline
BAM - CPTEC   & x           & 7.97 & 0.10 & 0.00 & 1.00 \\\hline
\end{tabular}
\label{tab:summer}
\end{table}

The LSTM model stood out as the best option, presenting an MSE of 1.28 and an R² of 0.85, which indicates a good predictive capacity. Furthermore, the POD of 0.59 and FAR of 0.35 show that it has an acceptable balance between accuracy in detecting heavy precipitation and reducing false alarms, although there is room for improvement in FAR. The latency of 2,736.7680 ms is also reasonable, which means that the model offers a good balance between accuracy and computational efficiency, making it an ideal choice for applications that require good performance without excessive latencies.

The CNN 1D and GRU showed similar performances, with MSE of 1.39 and 1.41, respectively, and R² of 0.84, but both had high latencies, which compromised their efficiency. Random Forest performed worse than LSTM (MSE of 1.45 and R² of 0.83) and had a latency similar to that of the GRU, making it less robust in terms of prediction. XGBoost is extremely fast compared to the other ML/DL models, latency of 3.6967 ms, but it had the worst accuracy (MSE of 1.53), making it the ML/DL model with the lowest accuracy in precipitation predictions, though it had reduced computational cost. However, it excelled in detecting intense precipitation events, as indicated by its more favorable POD and FAR values. BAM - CPTEC proved unfeasible, with MSE of 7.97 and R² of 0.10. Thus, LSTM stands out as the best choice in terms of prediction accuracy, making it the most suitable model when MSE is the priority. XGBoost, with its reduced computational cost, provided reasonable results for detecting intense events but had lower overall accuracy in MSE.

% Figure \ref{fig:summer-errorprec} presents the error map for LSTM, the best model for the summer season. Conseguimos notar as areas que o modelo subestimou e superestimou em termos da intensidade de precipitação prevista.

Figure \ref{fig:summer-errorprec} presents the error map for the LSTM, the best-performing model for the summer season. The red tones indicate areas where the model underestimated the precipitation, while the blue tones highlight regions where it overestimated the predicted intensity. These patterns allow clear identification of the model’s spatial biases.

\subsection{Autumn}

The autumn season in the Southern Hemisphere marks a transition between summer and winter. One of the characteristics of this season is the occurrence of persistent rains in the north and northeast regions of Brazil, often influenced by the Intertropical Convergence Zone (ITCZ), as stated in the autumn climate forecast \citep{inmetinpe2024prognostico}.

\begin{figure}[H]
    \centering
    \includegraphics[width=1.0\linewidth]{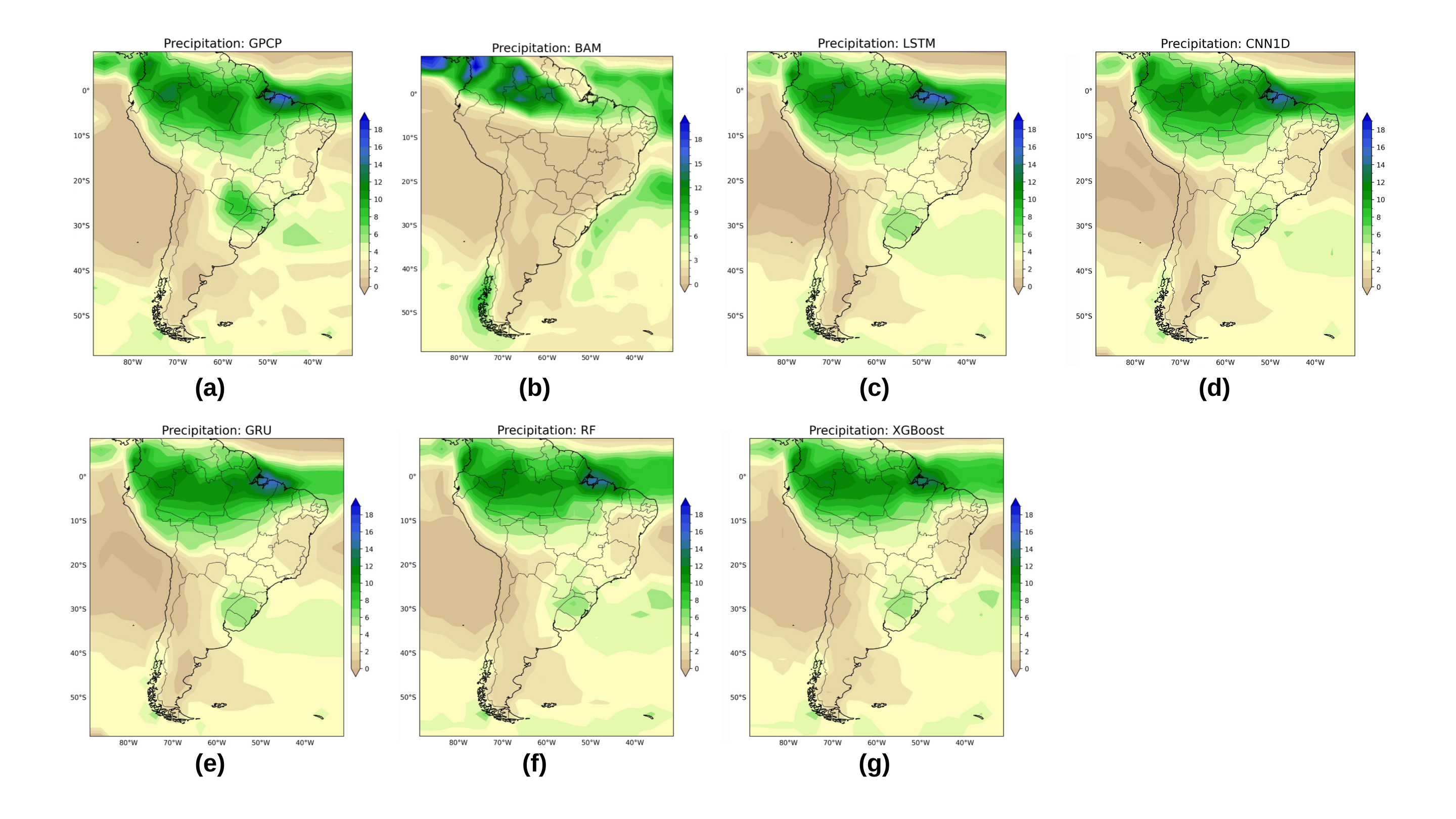}
    \caption{Maps of observed and predicted precipitation for the autumn of 2019. (a) precipitation observed by GPCP, (b) precipitation by the BAM model, (c) precipitation estimated by the LSTM model, (d) precipitation estimated by the CNN1D model, (e) precipitation estimated by the GRU model, (f) precipitation estimated by the RF model and (g) precipitation estimated by the XGBoost model.}
    \label{fig:autumn-mapsprec}
\end{figure}

\begin{figure}[H]
    \centering
    \includegraphics[width=0.4\linewidth]{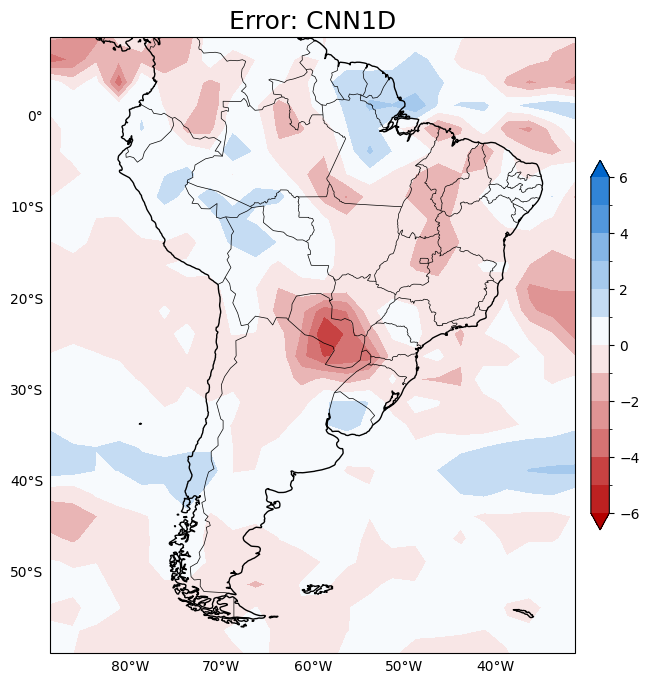}
    \caption{Error map of the best model for autumn season}
    \label{fig:autumn-errorprec}
\end{figure}

Figure \ref{fig:autumn-mapsprec} presents maps of precipitation observed by GPCP and precipitation predicted by each model during autumn 2019. Table \ref{tab:autumn} summarizes the results of the evaluation metrics for each model, while Figure \ref{fig:autumn-errorprec} highlights the error map associated with the best performing model.

In comparison with the GPCP observation shown in Figure \ref{fig:autumn-mapsprec}, the BAM model (b) presents a displacement of intense precipitation towards the extreme north of South America, in addition to overestimating the rainfall in this region and in the south of the continent. The deep learning models, LSTM (c), CNN1D (d) and GRU (e), capture the precipitation distribution well, but they smooth out some areas and may underestimate the maximum intensity observed in the GPCP. The models based on decision trees, RF (f) and XGBoost (g), present a similar pattern, but with a tendency to reduce extremes, especially in the Amazon and coastal regions. An important point is the difference in rainfall intensity in northern Brazil and the Andean region, where BAM shows higher values than other models. Furthermore, the transition between wet and dry areas is better defined in GPCP compared to machine learning models, which smooth out this variation.

\begin{table}[H]
\centering
\caption{Results of the models with what was observed in the GPCP. Best values for each metric are in bold.}
\vspace{0.2cm}
\begin{tabular}{|l| l| l| l| l| l|}\hline

Model & Latency (ms) & MSE & R$^2$ & POD & FAR \\\hline
CNN 1D        & 7,171.4450  & \textbf{0.87}  & \textbf{0.90} & 0.76 & 0.26 \\\hline
LSTM          & 2,843.0996  & 0.91  & 0.89 & \textbf{0.79} & \textbf{0.25} \\\hline
GRU           & 3,170.1417  & 1.05  & 0.88 & 0.71 & \textbf{0.25} \\\hline
Random Forest & 2,545.6480  & 1.14  & 0.87 & 0.68 & 0.34 \\\hline
XGBoost       & \textbf{3.4795}      & 1.17  & 0.86 & 0.65 & 0.31 \\\hline
BAM - CPTEC   & x           & 8.94  & -0.04 & 0.35 & 0.65 \\\hline

\end{tabular}
\label{tab:autumn}
\end{table}

CNN 1D was the most balanced model in the autumn, with MSE of 0.87, R² of 0.90 and a good balance between POD (0.76) and FAR (0.26). Despite the high latency (7,171.4450 ms), its predictive performance justifies the computational cost in this season. The LSTM achieved an MSE of 0.91, an R² of 0.89, and the highest POD (0.79), but its high latency (2,843.0996 ms) compromises efficiency. Moreover, when considering the remaining metrics, even though the LSTM has lower latency than the 1D-CNN, its forecasts were not more accurate than those produced by the 1D-CNN. Nevertheless, the results indicate an overall near-tie in performance between the 1D-CNN and the LSTM, making the LSTM one of the top-performing models for this season. GRU had the worst MSE (1.05) and R² of 0.88, in addition to an even higher latency (3,170.1417 ms). Random Forests and XGBoost performed worse (MSE of 1.14 and 1.17, respectively), with lower predictive capacity. The BAM - CPTEC was unfeasible (MSE of 8.94, R² of -0.04). Thus, CNN 1D stands out as the best option, balancing accuracy and computational efficiency.

% CNN 1D was the most balanced model in the autumn, with MSE of 0.87, R² of 0.90 and a good balance between POD (0.76) and FAR (0.26). Despite the high latency (7,171.4450 ms), its predictive performance justifies the computational cost in this season. LSTM had an MSE of 0.91, R² of 0.89 and the highest POD (0.79), but its high latency (2,843.0996 ms) compromises efficiency comparando com os resultados obtidos pela CNN1D. GRU had the worst MSE (1.05) and R² of 0.88, in addition to an even higher latency (3,170.1417 ms). Random Forests and XGBoost performed worse (MSE of 1.14 and 1.17, respectively), with lower predictive capacity. The BAM - CPTEC was unfeasible (MSE of 8.94, R² of -0.04). Thus, 1D CNN stands out as the best option, balancing accuracy and computational efficiency.

Figure \ref{fig:autumn-errorprec} presents the error map for the CNN 1D, the best-performing model for the autumn season. The red tones indicate areas where the model underestimated the precipitation, while the blue tones highlight regions where it overestimated the predicted intensity. These patterns allow clear identification of the model’s spatial biases.

\subsection{Winter}

Winter in the Southern Hemisphere is characterized by low precipitation intensity in the Southeast, Central-West and parts of the North and Northeast regions of Brazil. On the other hand, in the extreme south of South America, precipitation volumes increase due to the influence of cold fronts, extratropical cyclones, as stated in the Winter Climate Prognosis \citep{inmetinpe2024prognostico}.

\begin{figure}[H]
    \centering
    \includegraphics[width=1.0\linewidth]{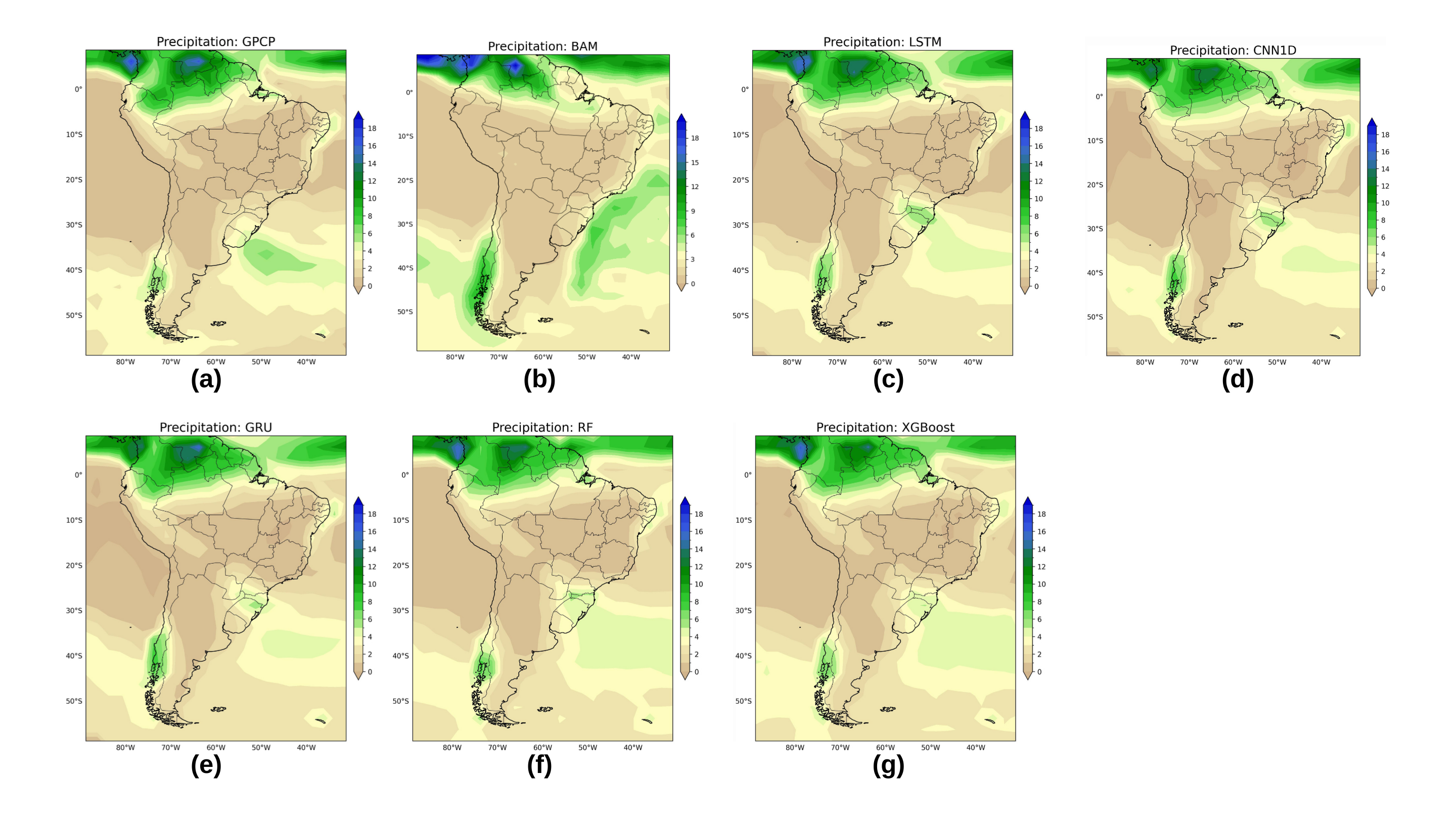}
    \caption{Maps of observed and predicted precipitation for the winter of 2019. (a) precipitation observed by GPCP, (b) precipitation by the BAM model, (c) precipitation estimated by the LSTM model, (d) precipitation estimated by the CNN1D model, (e) precipitation estimated by the GRU model, (f) precipitation estimated by the RF model and (g) precipitation estimated by the XGBoost model.}
    \label{fig:winter-mapsprec}
\end{figure}

\begin{figure}[H]
    \centering
    \includegraphics[width=0.4\linewidth]{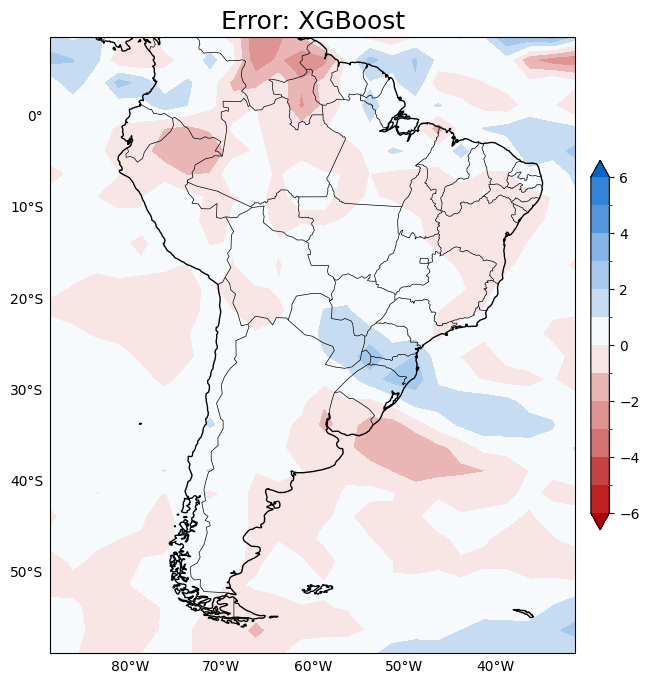}
    \caption{Error map of the best model for winter season}
    \label{fig:winter-errorprec}
\end{figure}

Figure \ref{fig:winter-mapsprec} presents the maps of precipitation observed by the GPCP and the precipitation predicted by each model during the winter of 2019. Table \ref{tab:winter} summarizes the results of the evaluation metrics for each model, while Figure \ref{fig:winter-errorprec} highlights the error map associated with the best performing model.

In winter, as shown in Figure \ref{fig:winter-mapsprec}, GPCP (a) shows the concentration of precipitation in the extreme north of South America and in the southern region of the continent. The BAM model (b) intensifies precipitation in the north, presenting higher values than the GPCP, and also suggests heavier rains in the southern region, especially in Argentina and Chile. The deep learning models, LSTM (c), CNN1D (d) and GRU (e), follow a similar pattern to GPCP, but smooth the precipitation intensity and present a more diffuse transition between dry and wet areas. The models based on decision trees, RF (f) and XGBoost (g), show more homogeneous precipitation patterns, but with less spatial variation, which may indicate a lower capacity to capture extremes. An important point is that the Amazon region appears less rainy in machine learning-based models compared to GPCP and BAM.

\begin{table}[H]
\centering
\caption{Results of the models with what was observed in the GPCP. Best values for each metric are in bold.}
\vspace{0.2cm}
\begin{tabular}{|l| l| l| l| l| l|}\hline

Model & Latency (ms) & MSE & R$^2$ & POD & FAR \\\hline

CNN 1D        & 7,130.8350  & 0.84 & 0.89 & 0.74 & \textbf{0.11} \\\hline
LSTM          & 2,826.6041  & 0.83 & 0.89 & 0.79 & 0.21 \\\hline
GRU           & 2,967.3042  & 0.75 & 0.90 & \textbf{0.85} & 0.15 \\\hline
Random Forest & 2,462.1010  & 0.75 & 0.90 & 0.79 & 0.25 \\\hline
XGBoost       & \textbf{4.2348}      & \textbf{0.62} & \textbf{0.92} & 0.82 & 0.22 \\\hline
BAM - CPTEC   & x           & 4.82 & 0.37 & 0.74 & 0.46 \\\hline

\end{tabular}
\label{tab:winter}
\end{table}

XGBoost was the best model in winter, achieving the lowest MSE (0.62) and the highest $R^2$ (0.92), indicating excellent prediction accuracy and good explanatory power of the data variability, along with the lowest latency (4.2348 ms), making it the most computationally efficient. While it showed a POD of 0.82 and FAR of 0.22, indicating slightly lower effectiveness in detecting heavy precipitation, its overall performance was superior. GRU performed well in terms of POD (0.85) and FAR (0.15), but with higher latency (2,967.3042 ms), while 1D CNN and LSTM also performed well, but with lower efficiency and accuracy. Random Forest had similar performance to GRU, but with a higher FAR, affecting its accuracy. BAM - CPTEC performed the worst.

Figure \ref{fig:winter-errorprec} presents the error map for the XGBoost, the best-performing model for the winter season. The red tones indicate areas where the model underestimated the precipitation, while the blue tones highlight regions where it overestimated the predicted intensity. These patterns allow clear identification of the model’s spatial biases.

\subsection{Spring}

Spring is a transition season between dry and rainy seasons in the Southern Hemisphere. In the extreme north of the continent there is intense precipitation and in the south there is more moderate precipitation \citep{inmetinpe2024prognostico}.

\begin{figure}[H]
    \centering
    \includegraphics[width=1.0\linewidth]{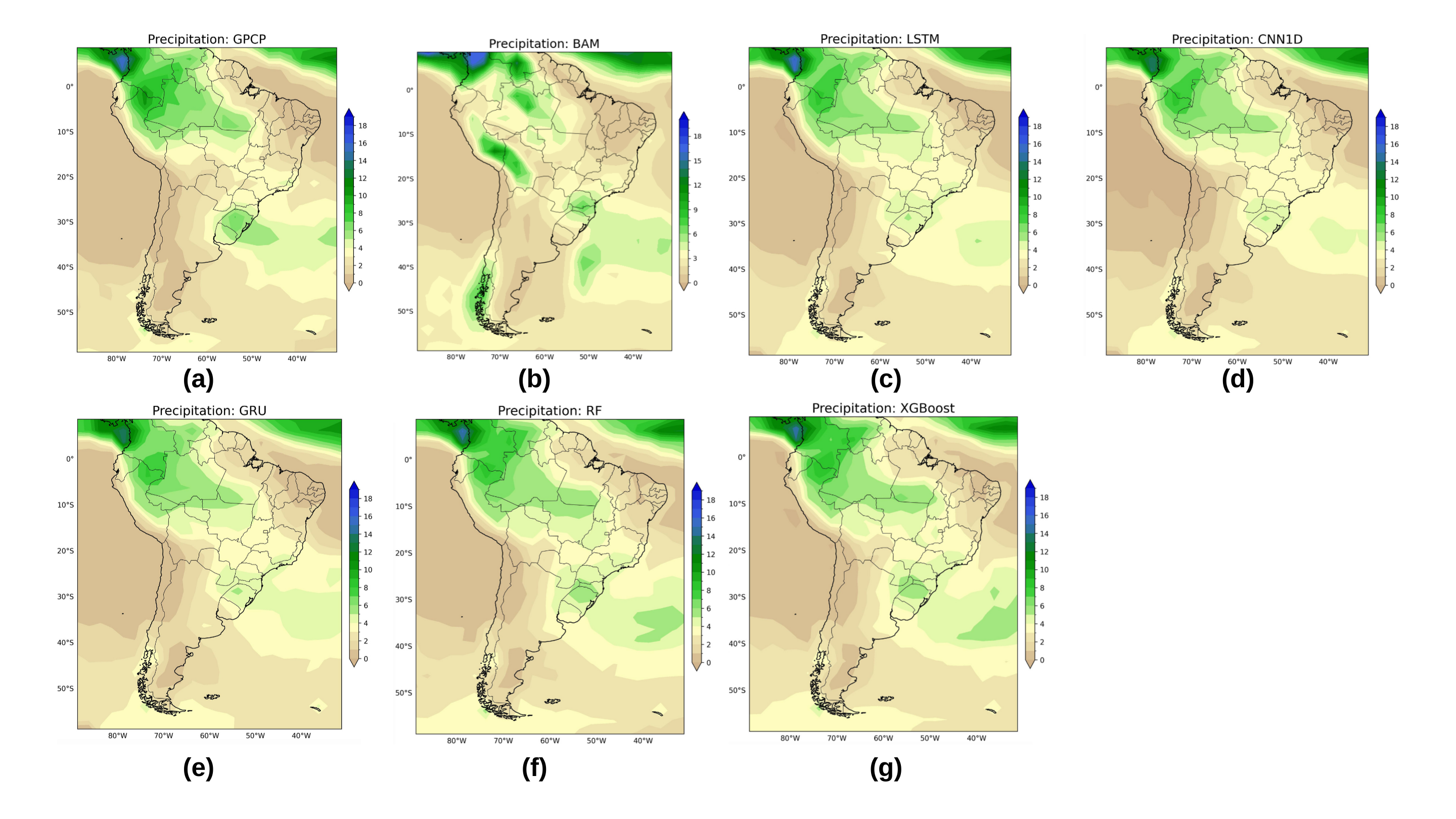}
    \caption{Maps of observed and predicted precipitation for the spring of 2019. (a) precipitation observed by GPCP, (b) precipitation by the BAM model, (c) precipitation estimated by the LSTM model, (d) precipitation estimated by the CNN1D model, (e) precipitation estimated by the GRU model, (f) precipitation estimated by the RF model and (g) precipitation estimated by the XGBoost model.}
    \label{fig:spring-mapsprec}
\end{figure}

\begin{figure}[H]
    \centering
    \includegraphics[width=0.4\linewidth]{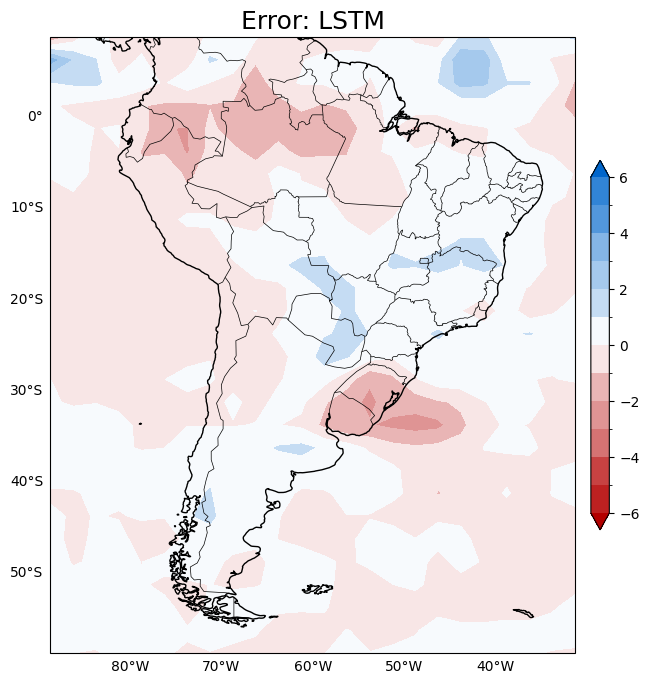}
    \caption{Error map of the best model for spring season}
    \label{fig:spring-errorprec}
\end{figure}

Figure \ref{fig:spring-mapsprec} presents maps of precipitation observed by GPCP and precipitation predicted by each model during spring 2019. Table \ref{tab:spring} summarizes the results of the evaluation metrics for each model, while Figure \ref{fig:spring-errorprec} highlights the error map associated with the best performing model.

In comparison with the GPCP observation (a) shown in Figure \ref{fig:spring-mapsprec}, it reveals that all models capture the general distribution of precipitation, but with variations in the intensity and location of wet and dry areas. The BAM model (b) presents a similar pattern to the GPCP, but with differences in precipitation intensity, especially in the Amazon region and southern Brazil. Deep learning models such as LSTM (c), CNN1D (d) and GRU (e) show good agreement in capturing precipitation in the Amazon and the Atlantic coast of Brazil, but tend to smooth out rainfall intensity. The tree-based models, RF (f) and XGBoost (g), reproduce the general structure of precipitation, however with discrepancies in the Chaco region, in central Argentina, and in the Northeast region of Brazil.

\begin{table}[H]
\centering
\caption{Results of the models with what was observed in the GPCP. Best values for each metric are in bold.}
\vspace{0.2cm}
\begin{tabular}{|l| l| l| l| l| l|}\hline

Model & Latency (ms) & MSE & R$^2$ & POD & FAR \\\hline

CNN 1D        & 7,347.1070  & 0.47 & 0.92 & 0.68 & \textbf{0.08} \\\hline
LSTM          & 2,765.9340  & \textbf{0.43} & \textbf{0.93} & 0.65 & \textbf{0.08} \\\hline
GRU           & 2,938.6680  & 0.44 & \textbf{0.93} & 0.62 & 0.09 \\\hline
Random Forest & 256.5870    & 0.44 & \textbf{0.93} & 0.59 & 0.09 \\\hline
XGBoost       & \textbf{3.2883}      & 0.48 & 0.92 & 0.68 & 0.12 \\\hline
BAM - CPTEC   & x           & 3.43  & 0.43 & \textbf{0.74} & 0.32 \\\hline

\end{tabular}
\label{tab:spring}
\end{table}

LSTM was the best model in the spring, with MSE of 0.43, R² of 0.93, good POD (0.65) and low FAR (0.08), ensuring high accuracy and few false detections. CNN 1D (MSE of 0.47, R² of 0.92) performed close but inferior. GRU (MSE of 0.44, R² of 0.93) had lower latency (2,938.668 ms), but a lower POD (0.62). Random Forests had MSE and R² equivalent to GRU, but the lowest POD (0.59), although with the lowest latency (256.587 ms). XGBoost (MSE of 0.48, R² of 0.92) was efficient in processing time (3.2883 ms), but had a higher FAR (0.12). BAM - CPTEC performed very poorly (MSE of 3.43, R² of 0.43). In conclusion, LSTM was the best model, balancing accuracy and precipitation detection.

Figure \ref{fig:spring-errorprec} presents the error map for the LSTM, the best-performing model for the spring season. The red tones indicate areas where the model underestimated the precipitation, while the blue tones highlight regions where it overestimated the predicted intensity. These patterns allow clear identification of the model’s spatial biases.

\subsection{Explainable AI Analysis for Autumn and Spring}

\begin{figure}[H]
    \centering
    \includegraphics[width=1.1\linewidth]{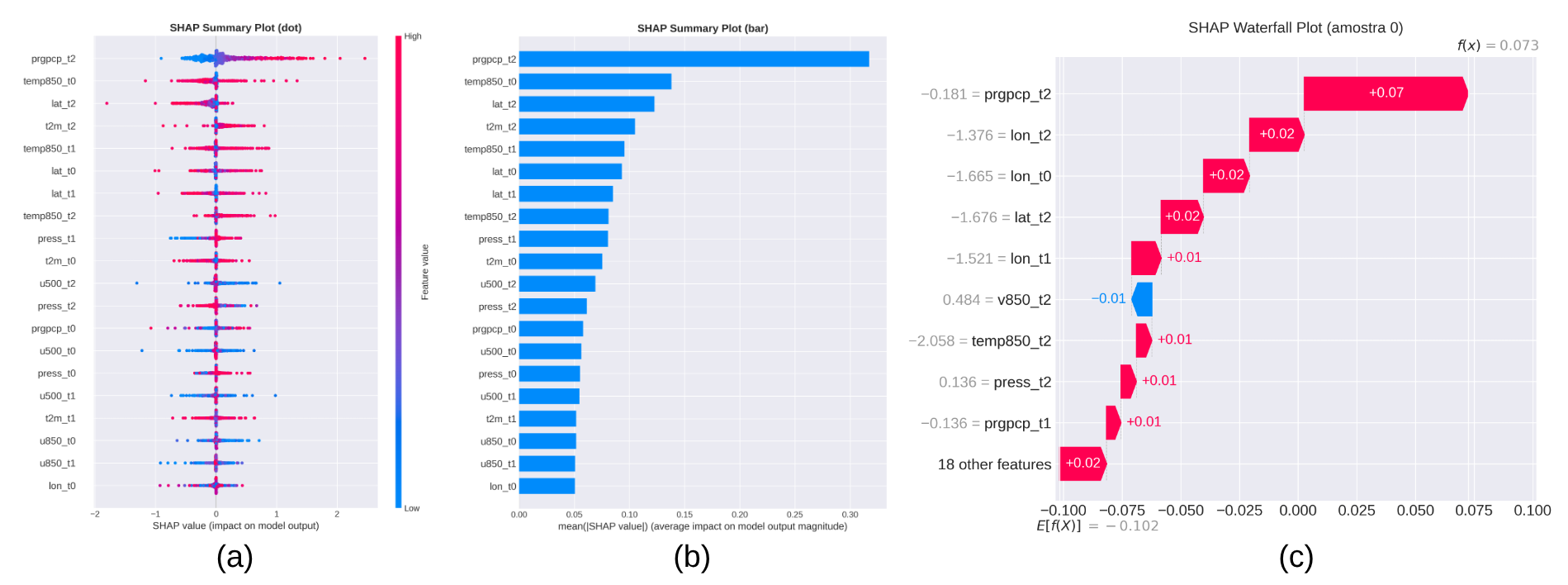}
    \caption{SHAP visualizations for the autumn season: (a) summary plot (dot), (b) summary plot (bar), and (c) waterfall plot.}
    \label{fig:autumn-shap}
\end{figure}

\begin{figure}[H]
    \centering
    \includegraphics[width=1.1\linewidth]{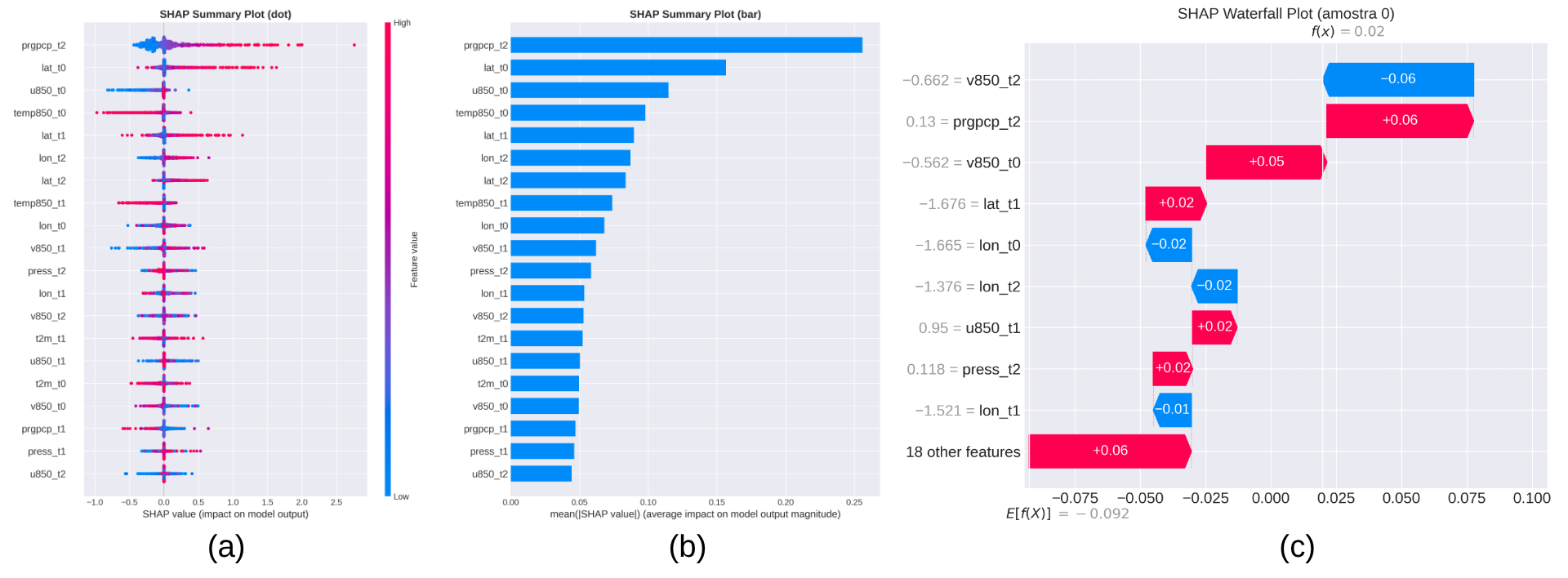}
    \caption{SHAP visualizations for the spring season: (a) summary plot (dot), (b) summary plot (bar), and (c) waterfall plot.}
    \label{fig:spring-shap}
\end{figure}

In this section, the SHAP value visualizations (Figures \ref{fig:autumn-shap} and \ref{fig:spring-shap}) corresponding to the transition seasons, autumn and spring, are presented together with a descriptive analysis of the results. The focus is placed on the LSTM model, which demonstrated superior performance overall in terms of predictive accuracy and the ability to detect intense precipitation events across all seasons. This analysis aims to enhance the explainability of the model by identifying the most influential variables contributing to precipitation prediction during these periods.

%interpretability of the model by identifying the most influential variables contributing to precipitation prediction during these periods.

Figure \ref{fig:autumn-shap} presents the explainability analysis based on SHAP values applied to the LSTM model trained to predict precipitation in the autumn season. In this model, each meteorological variable is represented at three consecutive instants (t0, t1, and t2), corresponding to a timestep $=$ 3.

In the summary plot presented in Figure \ref{fig:autumn-shap}(a), it can be observed that prgpcp\_t2 (precipitation lagged by two time steps) and temp850\_t0 (temperature at the 850 hPa level at the current instant) are the variables with the greatest impact on the forecasts. This indicates that the model heavily uses the temporal memory of precipitation and the thermodynamic state of the atmosphere to infer future seasonal behavior.

The summary bar plot presented in Figure \ref{fig:autumn-shap}(b) confirms that the absolute average importance of the variables is concentrated in the components associated with past precipitation, temperature, and spatial position (latitude and longitude). This result demonstrates that the model has learned a consistent spatiotemporal representation, using both the geographic context and antecedent conditions to infer the autumn rainfall response.

The waterfall plot presented in Figure \ref{fig:autumn-shap}(c) illustrates the local decomposition of an individual forecast, showing how the model adjusts the final value from additive sums of the effects of the variables. Variables with positive SHAP values (in red) increase the precipitation estimate, while negative values (in blue) reduce it. This decomposition shows that the network is making interpretable and consistent decisions, associating specific atmospheric states with seasonal rainfall responses.

In Figure \ref{fig:spring-shap}, we present the SHAP visualizations corresponding to the spring season.

Figure \ref{fig:spring-shap} presents the interpretability analysis based on SHAP values for the LSTM model trained to predict average precipitation for the spring season. The model uses timestep $=$ 3, so that each climate variable is considered at three consecutive time points (t0, t1, and t2).

In the summary plot presented in Figure \ref{fig:spring-shap}(a), it can be seen that prgpcp\_t2 (precipitation lagged by two time steps) has the greatest impact on the forecasts, followed by lat\_t0, u850\_t0 (zonal component of the wind at 850 hPa at the current time), and temp850\_t0 (temperature at 850 hPa at the current time). In terms of machine learning, this behavior suggests that the network is combining persistent patterns (past precipitation) and dynamic patterns (winds and temperature).

The summary bar plot presented in Figure \ref{fig:spring-shap}(b) reinforces this interpretation, showing that the most relevant variables are distributed across different time horizons, indicating that the network is adequately exploiting the sequential context

The waterfall plot presented in Figure \ref{fig:spring-shap}(c) details the local decomposition of a specific forecast. In it, we observe that variables such as v850\_t2 and v850\_t0 (meridional wind component) have a negative effect on the forecast, while prgpcp\_t2, u850\_t1, and press\_t2 (surface pressure at time t2) contribute positively.

\section{Discussion}

Analysis of the results reveals that the choice of the ideal model depends on different factors, such as predictive capacity, detection of intense precipitation and computational efficiency.

In the summer, the LSTM model stood out, presenting the best balance between MSE (1.28) and R² (0.85), which demonstrates a good predictive capacity. Furthermore, it had a reasonable POD (0.59) and a FAR of 0.35, which indicates a good performance in identifying precipitation, although still with a reasonable rate of false alarms. XGBoost, despite being the fastest model, was unable to achieve the accuracy of other models, presenting the highest MSE (1.53), which compromised its prediction performance.

In autumn, CNN 1D stood out again, with the best MSE (0.87) and R² (0.90). Furthermore, the POD of 0.76 and FAR of 0.26 indicate that this model has a good performance in detecting precipitation, with a relatively low false alarm rate. Although the CNN 1D delivered good results, it did not surpass the LSTM in predicting heavy rainfall, as the latter achieved a POD of 0.79 and a FAR of 0.25. Therefore, the LSTM can also be considered a suitable model for this season. The XGBoost was fast, but the lower POD and higher FAR didn't make it the best choice.

In winter, the XGBoost model stood out as the most efficient, achieving the lowest MSE (0.62) and the highest R² (0.92), reflecting excellent predictive accuracy and good explanatory power of the data variability. Additionally, with the lowest latency (4.2348 ms), it was the most computationally efficient model. Although its POD of 0.82 and FAR of 0.22 indicate slightly lower effectiveness in detecting heavy precipitation, its overall performance was superior. The GRU model performed excellently in terms of POD (0.85) and FAR (0.15), demonstrating a strong ability to identify heavy precipitation with a low false alarm rate. However, its higher latency (2,967.3042 ms) reduced its efficiency compared to XGBoost. Models like CNN 1D and LSTM also showed good results but had higher latency and slightly lower accuracy than XGBoost. Random Forest performed similarly to GRU, but with a higher FAR (0.25), which affected its accuracy, making it less robust. The BAM - CPTEC model performed the worst, with MSE of 4.82 and R² of 0.37, rendering it unfeasible for precipitation forecasting in winter. Therefore, XGBoost emerged as the most suitable model, balancing accuracy, explanatory power, and computational efficiency.

In spring, LSTM stood out, with the lowest MSE (0.43) and highest R² (0.93), reflecting excellent predictive capacity. The POD of 0.65 and the FAR of 0.08 indicate good performance in identifying precipitation, with a very low rate of false alarms. 1D CNN performed well, but LSTM was superior in terms of accuracy. XGBoost, although fast, had a higher MSE (0.48) and a higher FAR (0.12), which compromised its performance.

Across all seasons, LSTM was consistently one of the best models, with a good combination of low latency and high predictive power. XGBoost, although very efficient in terms of latency, was not as accurate, especially in detecting heavy precipitation. GRU stood out in the winter season, while Random Forests and CNN 1D performed well, but with some limitations in relation to the best models.

In summary, the choice of the ideal model depends on the focus of the application, whether in terms of accuracy, identification of intense precipitation or computational efficiency. LSTM balanced accuracy and detection capability well, while XGBoost and Random Forests were more efficient in terms of latency. So, if the focus is on accuracy and intense precipitation detection, LSTM would be the best option according to the presented results. However, if the computational cost is the determining factor, where accuracy and identification of intense precipitation are tolerable at lower values, eventually XGBoost could be the best choice.

In addition to the quantitative evaluation of model performance, SHAP-based analysis provided deeper insights into the explainability of the LSTM model, particularly regarding how different climate variables influenced precipitation forecasting. SHAP values revealed that variables associated with specific components of humidity, air temperature, and wind were the most relevant contributors across seasons, aligning with known physical mechanisms that drive precipitation. In the transitional seasons of autumn and spring, SHAP patterns indicated more intricate interactions between variables, reflecting the inherent atmospheric variability characteristic of these periods. Furthermore, variable importance analysis demonstrated that the temporal dependence between predictors plays a decisive role in the LSTM's ability to produce more accurate precipitation forecasts, including the correct identification of extreme precipitation events. These results reinforce the robustness of the LSTM architecture, highlighting not only its predictive accuracy but also its consistency with climatological principles, which strengthens confidence in its use for seasonal precipitation forecasting.

\section{Conclusion}

In this article, we aimed at addressing the need to perform broader investigations for forecasting precipitation in South America where artificial intelligence (ML/DL) models are used. The evaluated models showed varying performances across the seasons. LSTM stood out for its excellent predictive capability, with good accuracy, although it had higher latency, making it the best option when the focus is on precipitation forecasting and detecting intense events. GRU performed similarly to LSTM but also had higher latency. XGBoost, on the other hand, excelled in computational efficiency, with good latency, but compromised forecast accuracy, particularly for detecting intense precipitation, with inferior results in terms of MSE and POD. The CNN 1D performed well, but was surpassed by LSTM and GRU in terms of accuracy and the ability to detect temporal patterns. Random Forests also performed well but had higher latency and lower accuracy. BAM had the worst results.

Therefore, if the focus is on accuracy and intense precipitation detection, LSTM is the best option. However, if computational cost is the determining factor, with acceptable reductions in accuracy and intense precipitation detection, XGBoost could be the better choice due to its lower latency. For the future, we plan to use Optuna to optimize the hyperparameters, which may further improve model performance. Additionally, we aim to explore graph neural networks for forecasting in the same region of interest (South America), taking advantage of their benefits in capturing spatial and temporal dependencies, which could lead to better accuracy in climate predictions. We also plan to incorporate anomaly prediction to better understand unexpected variations and extreme events in the climate data, and to explore additional explainable AI techniques to further enhance the model’s ability to predict extreme precipitation events.

\section{Acknowledgements}

This research was developed within the project \textit{Classificação de imagens e dados via redes neurais profundas para múltiplos domínios} (Image and data classification via Deep neural networks for multiple domains - IDeepS). The IDeepS project (available online: \url{https://github.com/vsantjr/IDeepS}, accessed on 12 November 2025) is supported by the Laboratório Nacional de Computação Científica (LNCC - National Laboratory for Scientific Computing, MCTI, Brazil) via resources of the SDumont supercomputer. This research was carried out within the scope of the Laboratório de Inteligência ARtificial para Aplicações AeroEspaciais e Ambientais (Artificial Intelligence Laboratory for Aerospace and Environmental Applications - LIAREA) (available online: \url{https://liarealab.github.io/liarea_page/index.html}, accessed on 12 November 2025). This research was also supported by CAPES (Finance Code 88887.951221/2024-00).

\bibliographystyle{elsarticle-num-names} 
\bibliography{references}


% %% For authoryear reference style

% \bibitem[Lamport(1994)]{lamport94}
%   \textit{\LaTeX: a document preparation system},
%   Addison Wesley, Massachusetts,

% \end{thebibliography}
\end{document}